%% file: dpbn_tca.tex
\documentclass[conference]{IEEEtran}
\IEEEoverridecommandlockouts
\usepackage{cite}
\usepackage{amsmath,amssymb,amsfonts}
\usepackage{graphicx}
\usepackage{textcomp}
\usepackage{xcolor}
\def\BibTeX{{\rm B\kern-.05em{\sc i\kern-.025em b}\kern-.08em
    T\kern-.1667em\lower.7ex\hbox{E}\kern-.125emX}}

\usepackage{url}
\usepackage{amsfonts}

\title{Improved Auto-Encoding using Deterministic Projected Belief Networks
and Compound Activation Functions\thanks{This work was supported jointly by the Office of Naval Research Global and the Defense Advanced Research Projects Agency under Research Grant - N62909-21-1-2024}}

\author{\IEEEauthorblockN{Paul M Baggenstoss}
\IEEEauthorblockA{\textit{Fraunhofer FKIE} \\
        Fraunhoferstraße 20, 53343 Wachtberg, Germany\\
p.m.baggenstoss@ieee.org}
}

\begin{document}
\include{macros}
\newtheorem{identity}{Identity}
\newtheorem{hypothesis}{Hypothesis}
\newcommand{\mathtiny}[1]{\mbox{\tiny$#1$}}

\maketitle

\begin{abstract}
In this paper, we exploit the unique properties of a deterministic projected belief network (D-PBN)
to take full advantage of trainable compound activation functions (TCAs).
A D-PBN is a type of auto-encoder that operates by ``backing up" through
a feed-forward neural network.  TCAs are activation functions with complex 
monotonic-increasing shapes that change the distribution of the data 
so that the linear transformation
that follows is more effective.  Because a D-PBN operates by ``backing up",
the TCAs are inverted in the reconstruction process, 
restoring the original distribution of the data, thus taking
advantage of a given TCA in both analysis and reconstruction.  In this paper, we show
that a D-PBN auto-encoder with TCAs can significantly out-perform
standard auto-encoders including variational auto-encoders.
\end{abstract}
\vspace{-.0in}


\section{Introduction}
\subsection{Motivation}
Auto-encoders are an important type of generative network used in current tasks such as
unsupervised learning and open set classification \cite{Goodfellow2016},
and dimension-reduction is the main task involved.
It has been understood since the advent of support vector machines that 
non-linear operations can make the classification problem
linearly separable \cite{Sebald2001, SVMLight}. In the context
of auto-encoding, non-linear operations can make the distribution representable in a lower-dimensional
linear subspace. It is no surprise then, that the analysis task in auto-encoding
is implemented by alternating applications of non-linear transformation
(i.e. activation functions), followed by linear projection onto a lower-dimensional space.
While this concept is clear for the analysis task, it is fair to ask the
question: how can we construct an auto-encoder with non-linear transformations in such a way
that the return path (synthesis task) is more effective?
If non-linear transformations could be inverted during synthesis, then the
data distribution at the input would be restored.
The idea of inverting the non-linear transformation seems good at first,
because the data distribution is changed in the forward path,
and inverting it restores the original distribution.
But this makes less sense in an auto-encoder with separate analysis and synthesis
networks. But, a new type of auto-encoder based on the projected belief network (PBN)
has recently been proposed, in which the synthesis task is implemented by backing up through the
analysis network \cite{BagIcasspPBN,BagPBNHidim, BagPBNEUSIPCO2019, BagPBN}.
Therefore, inverting the non-linear transformation is inherently
part of the synthesis task.  We seek to exploit properties of the PBN together
with activation functions having complex shapes, to further improve the performance of an autoencoder.

\subsection{Main Concept}
A projected belief network (PBN) is unique among layered generative
networks because instead of using an explicit generative scheme,
it operates implicitly by backing-up through a feed-forward 
neural network (FFNN).  The FFNN serves as the optimal feature extractor \cite{BagIcasspPBN}, 
and is the dual network for the generative process \cite{BagPBN}.  
A PBN can also be sampled deterministically by choosing the conditional mean \cite{BagIcasspPBN},
resulting in the deterministic PBN (D-PBN).  
By passing forward through the FFNN, then backing
up using D-PBN, a type of auto-encoder is formed \cite{BagIcasspPBN,BagPBNHidim, BagPBNEUSIPCO2019, BagPBN}.
By working forward, and then backward, an auto-encoder based on D-PBN
is in a position to take full advantage of trainable compound activation functions (TCAs),
which are activation functions with complex monotonic-increasing shapes.
Like any non-linearity, TCAs can change the distribution of the data so that the linear transformation
that follows is more effective\cite{BagNNL}.  By inverting the TCAs in the sampling process
(synthesis task), the original distribution of the data is restored, realizing the advantage of a TCA 
in both analysis and reconstruction.

\subsection{Previous Work}
There is existing work on parameterizing activation functions,
but these are either of low complexity, or are not necessarily invertible
\cite{MolinaFlexible2020,he2015delving}.  Work on invertible activation 
functions exists in conjunction with normalizing flows (NF) \cite{NormFlows2021}, which is
a method related to PBNs that has recently gained popularity.
In NF, the network consists of a series of bijective (1:1) transformations, and the final output is 
usually regarded as being a set of independent identically-distributed ({\it iid}) random variables
(RVs).  The data generation process, is therefore the reverse of this: a set of {\it iid} RVs are subjected
to the inverse series of transformations. 
Activation functions with complex shapes based on splines have been proposed for NF
\cite{DurkanSpline2019,DurkanSpline2019b}. 
In NFs, however, there is no dimension reduction, so 
this body of work does not encompass auto-encoders.  
In order to get around the requirement
of dimension-preserving tansformations, a way to combine NF with variational autoencoders (VAEs)
has been described, \cite{Nielsen_SurVAE_NEURIPS2020}, however
this is a re-invention of PDF projection.

\subsection{Motivation to use TCAs}
\label{tcamotiv}
TCAs are activation functions with complex monotonic increasing shapes
that can make the linear transformation
to a lower-dimensional space more effective, reducing the dimension
required to represent the information. This is analogous to
the motivation for non-linear projections in support vector machines (SVMs) ,
making data linearly separable \cite{Sebald2001, SVMLight}.
A TCA  can transform data with modal clusters into data with a uni-modal
distribution if  the activation function approximates the cumulative
distribution of the input data \cite{BagNNL}.  Therefore,
 the first derivative of the transfer function should approximate the distribution of the input RV.
The usefulness of this property is obvious, especially when the operation
is inverted, allowing synthesizing multi-modal data from uni-modal data.

\subsection{Novelty and Contributions}
Both D-PBN and TCA have been previously separately introduced.
The main novelty and contribution of this paper is the
combination of D-PBN with TCA.  The advantage of using TCA in D-PBN is that the
TCA is inverted in the data synthesis process, thus
restoring the original data distribution, even if it is multi-modal.
Because D-PBN is the only kind of auto-encoder based on back-projection, it is the only
auto-encoder where a given TCA can be used to advantage in both
the analysis and reconstruction processes.

\section{Mathematical Approach}

\subsection{Review of PDF Projection}
The PBN and D-PBN are based on PDF projection, which we now review.
Subject to mild constraints, any fixed dimension-reducing 
transformation, $\bfy=T(\bfx)$, together with the known or assumed 
feature distribution $g(\bfy)$, corresponds to a probability density function (PDF) 
on the input data \cite{BagPDFProj,BagMaxEnt2018,Bag_info} given by
$G(\bfx) = \frac{p_{0,x}(\bfx)}{p_{0,x}(\bfy)} g(\bfy),$
where $p_{0,x}(\bfx)$ is a prior distribution and $p_{0,x}(\bfy)$
is its mapping to $\bfy$.  In the notation ``g" represents
the {\it given} feature distribution, ``G" is its
projection to the higher-dimensional range of $\bfx$, 
and subscript ``0" is a reminder that $p_{0,x}(\bfx)$ 
can be seen as a {\it a reference distribution}.  In our simplified notation,
the argument of a distribution defines its region of support, while
the subscript defines the space where the prior distribution
was originally applied.  Therefore $p_{0,x}(\bfy)$
has region of support on the range of $\bfy$, but is derived
from the reference distribution on $\bfx$.
On the other hand, $p_{0,y}(\bfy)$, which we will introduce below,
is a new reference distribution defined for $\bfy$.
Note that if $p_{0,x}(\bfx)$ is selected
for maximum entropy, then $G(\bfx)$ is unique.

The PDF of the input data can be estimated by training the
parameters of the transformation to maximize the mean of
$\log G(\bfx)$, resulting in a transformation
that extracts sufficient statistics and maximizes information \cite{BagKayInfo2022}.  
And, when $g(\bfy)$ is fixed, then the true distribution under the available 
training data will be driven toward $g(\bfy)$ \cite{BagKayInfo2022}.
We say that $G(\bfx)$ is the ``projection" of $g(\bfy)$ back to the input data.

To draw a sample of $G(\bfx)$, we first draw a sample $\bfy$ from
$g(\bfy)$, then draw $\bfx$ randomly from the set
$\{\bfx:T(\bfx)=\bfy\},$ weighted by $p_{0,x}(\bfx)$.
PDF projection admits a chain-rule by applying the idea recursively 
to stages of a transformation. 
Consider the cascade of two transformations $\bfy=T_1(\bfx)$, and $\bfz=T_2(\bfy)$.
Applying the above idea recursively, we have
$G(\bfx) = \frac{p_{0,x}(\bfx)}{p_{0,x}(\bfy)}  \frac{p_{0,y}(\bfy)}{p_{0,y}(\bfz)} g(\bfz).$
Note that  $p_{0,x}(\bfy)$ and  $p_{0,y}(\bfy)$ are two different
distributions with support on the range of $\bfy$. 
While $p_{0,x}(\bfy)=T_1[p_{0,x}(\bfx)]$,
 $p_{0,y}(\bfy)$  is a canonical MaxEnt prior distribution for $\bfy$.
The sampling process is also recursive, working backwards through the 
cascade of transformations.

\subsection{Review of D-PBN}
When PDF projection is applied layer by layer to a feed-forward neural network (FFNN), 
the result is a projected belief network (PBN) \cite{BagPBN}.  The PBN is stochastically 
sampled by ``backing up" through the FFNN, but
can be deterministically sampled by choosing the conditional mean estimate
in each layer, resulting in a deterministic PBN (D-PBN).
In any given layer of a FFNN, the hidden variable $\bfy\in\mathbb{R}^M$
is computed from the layer input $\bfx\in\mathbb{R}^N$, where
we assume $N>M$, i.e. a dimension-reducing layer.  Let 
$\bfy=\lambda\left({\bf b}+{\bf W}^\prime \bfx\right),$
where $\lambda(\;)$ is the element-wise activation function,
assumed to be strictly monotonic increasing (SMI).
In the D-PBN, we seek to reconstruct $\bfx$ from $\bfy$.
Because the SMI activation function and bias 
are invertible, we prefer to work with the output of the linear transformation, denoted by
$\bfz={\bf W}^\prime \bfx.$  Specifically, $\bfx$ is drawn randomly
 from the manifold ${\cal M}(\bfz) = \{ \bfx : {\bf W}^\prime \bfx = \bfz\},$
i.e.  we draw $\bfx$ from the set of samples that map to $\bfz$.
For the PBN, we draw $\bfx$ randomly from ${\cal M}(\bfz)$ proportional to  
a prior distribution $p_{0,x}(\bfx)$,
and for the  D-PBN, we select the conditional mean.  
We select $p_{0,x}(\bfx)$ according to the principle of maximum entropy \cite{Jaynes57} (MaxEnt).
Accordingly, $p_{0,x}(\bfx)$ has the highest entropy subject to any constraints 
which include the range of the variable $\bfx$, denoted
by $\mathbb{X}^N$, determined
by which activation function was used in the up-stream layer.
We consider three canonical ranges, $\mathbb{R}^N$ : $x_i\in (-\infty , \infty), \forall i$, $\mathbb{P}^N$ : $x_i\in [0 , \infty), \forall i$, and
 $\mathbb{U}^N$ : $x_i\in [0  , 1], \forall i$.
For a MaxEnt distribution to exist in the first two cases, we need constraints on the variance
\footnote{In the case of $\mathbb{P}^N$, a constraint on the mean is adequate, but this leads to
an uninteresting solution.}.

The canonical MaxEnt priors for
the ranges $\mathbb{R}^N$, $\mathbb{P}^N$, and $\mathbb{U}^N$,
are Gaussian, truncated Gaussian (TG), uniform,
respectively, which are listed in Table \ref{tab1a}. The uniform distribution is a special 
case of the truncated exponential distribution (TED). 
\begin{table}
\begin{center}
 \begin{tabular}{|l|l|l|}
\hline
         $\mathbb{X}^N$ &  MaxEnt Prior $p_{0,x}(\bfx)$  & $\lambda(\alpha)$ \\
 \hline
         $\mathbb{R}^N$   & $\prod_{i=1}^N {\cal N}(x_i)$  (Gaussian) & $\alpha$  (Linear)\\
 \hline
         $\mathbb{P}^N$   & $\prod_{i=1}^N 2 {\cal N}(x_i), \;\; 0<x_i $ (TG) & $\alpha + \frac{{\cal N}(\alpha)}{\Phi(\alpha)}$ (TG)\\
 \hline
         $\mathbb{U}^N$   & $1, \;\; 0<x_i<1$  (Uniform) & $\frac{e^{\alpha}}{e^{\alpha} - 1}-\frac{1}{\alpha}$ (TED)  \\
 \hline
\end{tabular}
\end{center}
\caption{MaxEnt priors and activation functions as a function of input data range.
        TG=``Trunc. Gauss.". TED=``Trunc. Expon. Distr".
${\cal N}\left(x\right) \defined \frac{e^{-x^2/2}}{\sqrt{2\pi}}$ and $\Phi\left( x\right)  \defined \int_{-\infty}^x {\cal N}\left(x\right).$
}
\label{tab1a}
\end{table}
Additional mathematical details can be found in the references \cite{BagIcasspPBN,BagPBNHidim}.
As estimate of $\bfx$, we use the conditional mean 
 $\hat{\bfx}=\mathbb{E}_0(\bfx|\bfz)$, i.e. the 
conditional mean under prior $p_{0,x}(\bfx)$.  In \cite{BagIcasspPBN} ,
we show that this is given by
$\hat{\bfx}=\mathbb{E}_0(\bfx|\bfz) = \lambda\left( {\bf W} \bfh\right),$
where $\lambda\left(\;\right)$ is a special MaxEnt activation function
corresponding to the MaxEnt prior (listed in Table \ref{tab1a}), 
and $\bfh$ is the saddle point, which solves the equation
\beq
{\bf W}^\prime \lambda\left( {\bf W} \bfh\right) = \bfz.
\label{spaeq}
\eeq

Sampling a D-PBN, therefore consists of the following steps:
(a) Assume that hidden variable $\bfy$ is given.
  (b) Invert the activation function and bias to obtain the linear
	    transformation output variable $\bfz$.
    (c) Solve (\ref{spaeq}) to obtain $\bfh$.
     (d)  Let $\hat{\bfx}=\lambda\left( {\bf W} \bfh\right).$
     (e) Now set the variable $\bfy$ at the output of the
	     previous layer to $\hat{\bfx}$, and repeat for the
	      next layer (working backwards).
As an aside, it was pointed out in \cite{BagIcasspPBN,BagPBNEUSIPCO2019}
that the inversion of the activation function in step 2 can be avoided
if the FFNN uses MaxEnt activations in all layers (but not applicable when if TCAs are used).

{\bf D-PBN as Auto-Encoder.}
The D-PBN forms an auto-encoder if the starting feature is obtained by passing
input data through the FFNN.  In each layer,
solving (\ref{spaeq}) presents a computational challenge
due to the need to repeatedly invert matrices of the size $M\times M$, where
$M$ is the output dimension of a layer.  The problem can be 
avoided or mitigated by various methods \cite{BagPBNHidim}, 
some of which will be employed in the experiments.

{\bf Sampling failure.} 
The solution to (\ref{spaeq}) is not guaranteed to exist
unless $\bfz={\bf W}^\prime \bfx$ for some $\bfx$ in the
range of $\bfX$, denoted by $\mathbb{X}^N$ \cite{BagIcasspPBN}. When backing up through
a FFNN, this condition is only strictly met for the last layer
(first layer when working backwards), meaning that some input samples cannot be auto-encoded.
Luckily, the {\it sampling efficiency}, which is the fraction of samples that succeed, 
can be driven to 1.0 with proper initialization and training
\cite{BagIcasspPBN,BagPBNEUSIPCO2019,PBNTk}. 
Interestingly, a network can be initialized as a
stacked RBM to have a sampling efficiency of 1 using the
up-down algorithm, a multi-layer
version of the contrastive divergence (CD) algorithm
for fine-tuning stacked RBMs\cite{HintonDeep06}.
To use TCAs in the initial stacked RBM, the TCAs are initialized to 
a neutral state so that operates similar to 
a simple activation function (the base activation funcion, See Section \ref{tcasec}),
and simple activation functions are used in the return (synthesis) path. 


\subsection{Review of TCAs}
\label{tcasec}
In Section \ref{tcamotiv}, we provided a motivation for TCAs. We now define the TCA precicely.
Consider the compound activation function $f(x)$ given by
\beq
f(x) = \frac{ \sum_{k=1}^K  \; e^{a_k}\; f_k\left(e^{w_k} x+b_k\right)}
{\sum_{k=1}^K  \; e^{a_k}},
\label{tcadef}
\eeq
where the functions $f_k(x)$ are simple activation functions, and  ${\bf w}=\{w_k\}$, ${\bf a}=\{a_k\}$ and ${\bf b}=\{b_k\}$
are scale, weights, and bias parameters, respectively.  The use of exponential functions is to insure positivity
of the weights and scale factors. Note that if $f_k(x)$ are SMI, then the TCA is SMI.
As a convention, all activation functions except
the first activation function $f_1(x)$ are sigmoid-like activations (see TED 
activation in Table \ref{tab1a}). 
Because these are limited to the range $[0,1]$, the ultimate data range of the TCA depends
only on $f_1(x)$ which is called the {\it base activation}.
The TCA behaves like the base activaton, but with $K-1$ ``wiggles".
The form of (\ref{tcadef}) is motivated by approximation of distributions with mixtures \cite{BagNNL}.  
More to the point, (\ref{tcadef})  has a first derivative closely approximated by 
a Gaussian mixture \cite{BagNNL},
which is know to be a universal approximator of distributions \cite{Redner}.
It is the first derivative that is propagated as PDF estimate in a PBN \cite{BagKayInfo2022}.  


\subsection{D-PBN + TCA}
\begin{figure}[h!]
  \begin{center}
    \includegraphics[width=3.4in]{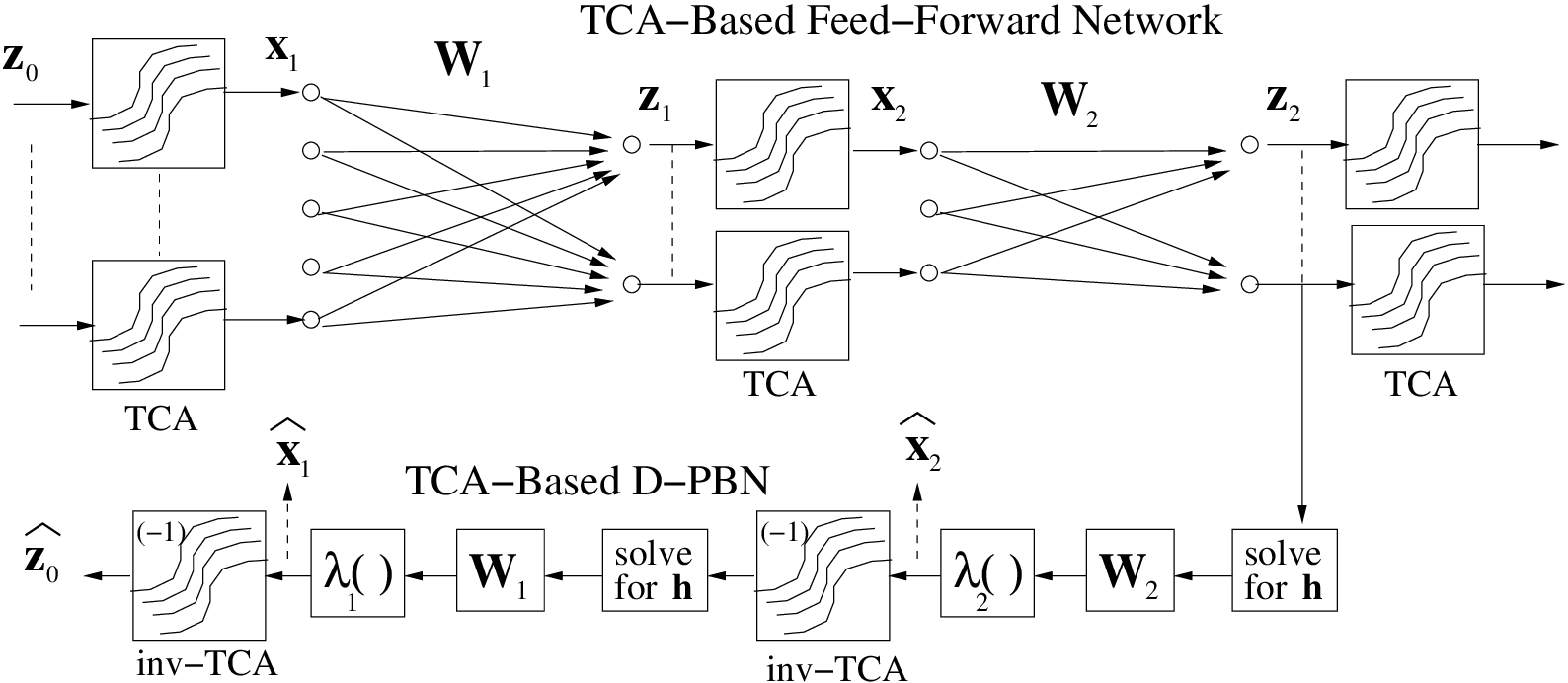}
  \caption{Block diagram of a 2-layer D-PBN auto-encoder employing TCAs.
	  Layer number is shown as subscript.}
  \label{pbn_nnl}
  \end{center}
\end{figure}

We propose to use TCAs in the FFNN forward path, and TCA inversion in the 
D-PBN reconstruction path.
A diagram of a D-PBN auto-encoder 
employing TCAs is shown in Figure \ref{pbn_nnl}.
No explicit bias is needed because it is included in the TCAs. 
Each D-PBN layer must choose one of the canonical data ranges,
MaxEnt priors, and activation functions (Table \ref{tab1}),
which in the case of the TCA is identified with the base activation.
Inversion of the TCA (i.e. find the $x$ for which $f(x)=y$)  
is done by bisection search, then a few Newton-Raphson iterations.  In a D-PBN, 
the last TCA in the forward path by itself does nothing, because
it gets immediately inverted in the sampling process.  

\section{Experiment 1}
\subsection{Data Set and Goal}
\label{ddesc1}
The the first data set was a subset of the  MNIST handwritten data corpus,
just three characters ``3", ``8", and ``9".
The data consists of sample images of $28\times 28$, or a data dimension of 784.
We used 500 training samples from each character, for a total of 1500 samples.
Since MNIST pixel data is coarsely quantized in the range [0,1],
a dither was applied to the pixel values\footnote{For pixel values
above 0.5, a small exponential-distributed random value was subtracted,
but for pixel values below 0.5, a similar  random value was added.}.
After applying dither, the data was ``gaussianified", i.e. transformed
from range [0, 1] to the real line by applying the logit (inverse sigmoid) function.
The resulting pixel values had a smooth amplitude distribution in the range -10 and 10.
Using gaussianified data made the use of mean-squared error more meaningful.
To facilitate duplicating the experimental results, 
the data we used is publicly available \cite{PBNTk}.
The goal is to auto-encode into a 16-dimensional bottleneck feature, then reconstruct the data with minimum
mean-squared error. Training was un-supervised (no label information used). 

\subsection{Network}
The network had three dense layers with 64, 32, and 16 nodes, respectively.
Both TCA and non-TCA networks were tested.
For the outputs of the first and second layer, we used 3-component
TCAs.  The base activation function is the TG (see Table \ref{tab1a}), which is very similar 
to softplus. We employed a 2-component TCA between the input data and first layer which
used the linear base activation.
A second, deeper network with 5 layers of 64, 48, 32, 24, and 16 nodes was also
tested for the standard auto-encoder to dispell the
notion that the TCA is nothing more than a type of additional layer.

\subsection{Algorithms and Training}
The following algorithms were tested:
{\bf AEC} : Standard auto-encoder with TG activation in all layers
and untied weights (decoder network is the transpose of the analysis network, but with independent weights).
{\bf VAE}: Variational auto-encoder. The VAE has a special output layer, so the final
	  TG activation is not used.
{\bf UPDN}:  Stacked-RBM fine-tuned using the up-down algorithm \cite{HintonDeep06}.
{\bf D-PBN}:  Deterministic PBN using TG activation functions.
{\bf D-PBN/TCA}: Deterministic PBN using TCAs with TG base activation functions
	    employing 3 components (base + 2).
For data augmentation we used random shifts of a maximum of 1 pixel in both vertical and horizontal
directions. Shifts were circular and calculated using the 2D FFT to attain vernier 
translation.

\subsection{Results}
\begin{table}
\begin{center}
 \begin{tabular}{|l|l|l|l|}
 \hline
	 Algorithm & Layers & MSE (train)  & MSE (test)  \\
 \hline
	 AEC &   3  & 3.05  & 3.15 \\
 \hline
	 AEC &   5  & 2.86  & 3.22 \\
 \hline
	 VAE &   3  & 3.87  & 4.18 \\
 \hline
	 UPDN &   3  & 3.41 & 3.46 \\
 \hline
	 D-PBN &   3  & 3.42 & 3.45 \\
 \hline
	 D-PBN/TCA &   3  & {\bf 2.84} & {\bf 2.96} \\
 \hline
\end{tabular}
\end{center}
\caption{Mean squared error (MSE) for Experiment 1.}
\label{tab1r}
\end{table}
As can be seen in Table \ref{tab1r}, the performance
of D-PBN/TCA was superior to all other
algorithms. The closest competitor 
for MSE on testing data was the 3-layer AEC,
which due to the independent decoder network, has
significantly more parameters.  We also
see with the 5-layer network that the
additional layers do not help, but rather
worsen overtraining. 

\section{Experiment 2}
\subsection{Data set and Goal}
Because MNIST data is widely seen to be over-simplified,
we conducted a second experiment using more realistic data
consisting of Google speech commands \cite{GoogleKW}.
We trained un-supervised on two difficult to distinguish words: ``three, and ``tree",
sampled at 16 kHz and segmented into
into 48 ms Hanning-weighted windows shifted by 16 ms.
We used log-MEL band energy features with 20 MEL-spaced
frequency bands and 45 time steps, representing a frequency span of 8 kHz and a time span of 0.72 seconds.
The input dimension was therefore $N=45\times 20=900.$
The number of training samples were 834, and 840, and testing samples were 430, and 449 for the two classes, respectively. The networks were trained unsupervised, taking all the data at once.
The goal is to auto-encode into a 32-dimensional feature, then reconstruct the data with minimum
mean-squared error. 

\subsection{Network and Training}
We used a network with two convolutional layers, 
followed by two dense layers of 64 and 32 neurons.
To mitigate the high computational load of 
D-PBN, the first 2 layers (convolutional) used linear activation,
forming a Gaussian group, resulting
in a much more efficient implementation of D-PBN \cite{BagPBNHidim}.
A TCA with 2 components and linear base activation
was applied to the input data,
TCAs with 3 components and TG base activation
were used before each dense layer.
For all non-D-PBN algorithms, the same network structure was used
with TG activation (similar to softplus) in all layers.

The PBN network (with or without TCA) was trained using the Theano framework
with PBN Toolkit \cite{PBNTk} 
with a mini-batch size of 288 samples and weight-decay.
No data augmentation was used.
Training a PBN with TCA requires using a very small learning rate
due to the weak and highly non-linear 
dependence on the parameters of the TCA.
Training took a week using a Quadro-P6000 GPU.

To give AEC the best chance, we asked a colleague to 
develop a Tensorflow/Keras-based implementation
with no constrains (gloves off).  The best-performing network,
listed as ``Unconstrained" in the results,  had 
21 layers and used ReLu activation
and batch normalization (our networks did not).   
Data and D-PBN code is available for reader verification
at \cite{PBNTk}.

\subsection{Results}
\begin{figure}[h!]
  \begin{center}
    \includegraphics[width=3.4in,height=1.8in]{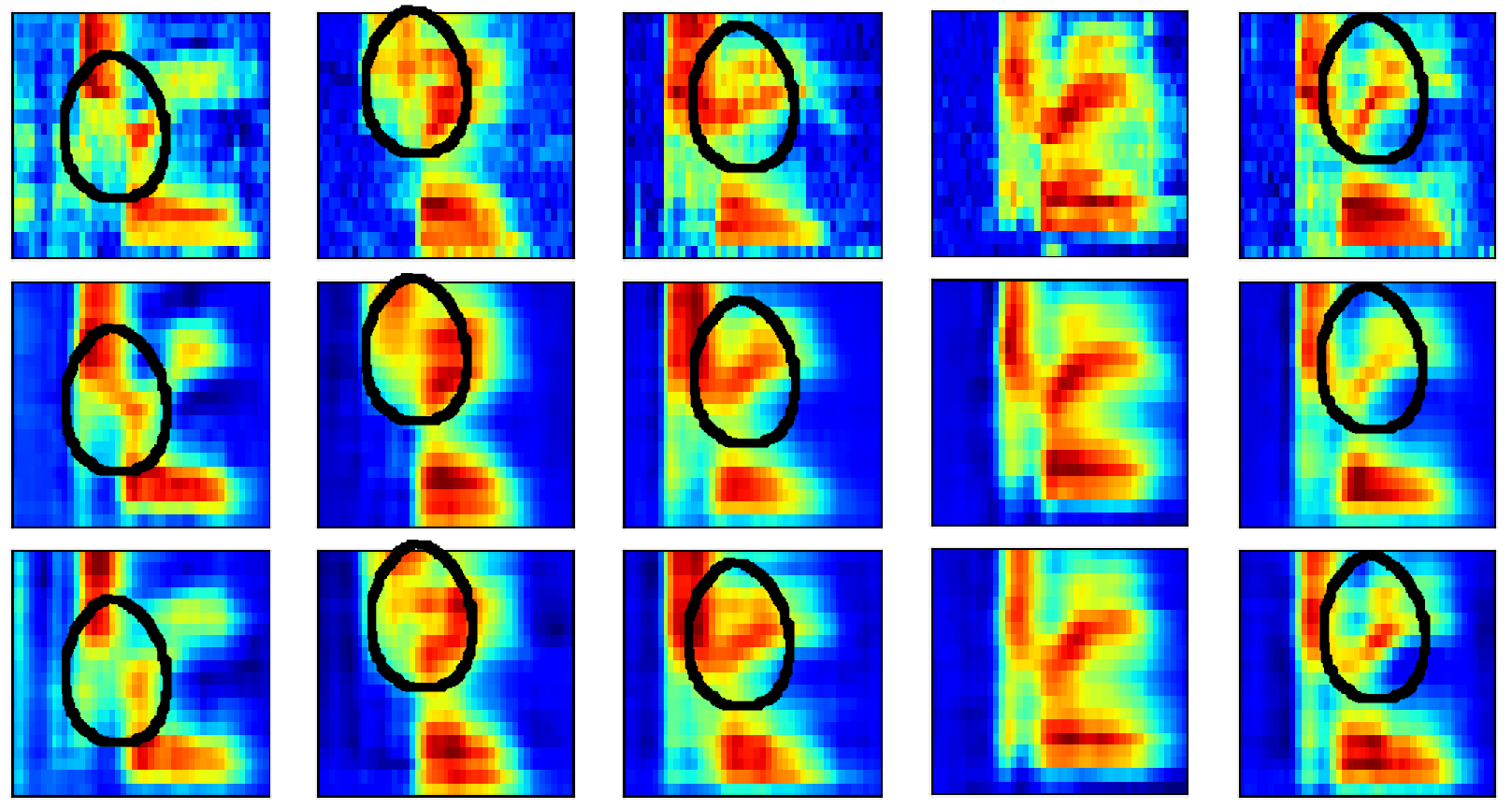}
  \caption{Comparison of reconstructions for testing data. Top row: original
spectrograms. Center: AEC (Untied).  Bottom: PBN/TCA. Examples of areas where PBN/TCA
shows better detail are circled. }
  \label{comp}
  \end{center}
\end{figure}

Results of Experiment 2 are provided in Table \ref{tab1}.
The validated mean squared error for D-PBN/TCA is significantly lower than all
other algorithms.
We provide examples of reconstructed spectrograms
in Figure \ref{comp} in which it is seen that D-PBN/TCA 
visibly preserves the details better than the best of the other algorithms.
We have outlined details in the Figure that show notable improvement.
Details for DPBN/TCA (bottom row) are closer to the original (top row)
that AEC (middle row).
\begin{table}[htb]
\begin{center}
 \begin{tabular}{|l|l|l|l|}
\hline
	 Algorithm &  L2 reg  & MSE (Train) & MSE (Test) \\
 \hline
	 AEC/Tied (*)   &   1e-3 & 2.500  & 3.093(**)  \\
 \hline
	 AEC/Untied (*)   &   1e-3 &  2.405  & 3.0558(**)  \\
 \hline
	 Unconstrained  &  0 &    {\bf 2.38}  & 3.02 \\
 \hline
	 PBN       &  0 &  2.777        & 3.030  \\
 \hline
	 PBN/TCA   &  0 &  2.4102        & {\bf 2.8092}  \\ 
 \hline
\end{tabular}
\end{center}
	\caption{Mean squared reconstruction error (MSE) for various algorithms.
	(*) Tied: the analysis and reconstruction weights are the same except
	  for a trainable scale factor. Untied: weights are independent.  
	   (**) Optimized over L2 regularization factor.  }
        \label{tab1}
\end{table}


\section{Conclusions and Future Work}
In this paper, we demonstrated that TCAs and D-PBN complement each other
so that together, they to provide a significant advantage in validated mean-squared error
over other types of auto-encoders.  Future work is needed to make the training more efficient.
The weak and highly non-linear dependence on the TCA parameters might be mitigated
by re-parameterizing the TCAs.
Data, software, and parameters to recreate these results are provided at \cite{PBNTk}.

\bibliographystyle{ieeetr}
\bibliography{ppt}
\end{document}

%% file: macros.tex
\newcommand{\defined}{\stackrel{\mbox{\tiny$\Delta$}}{=}}
\newtheorem{example}{Example}
\newtheorem{conclusion}{Conclusion}
\newtheorem{assumption}{Assumption}
\newtheorem{definition}{Definition}
\newtheorem{problem}{Problem}
\newcommand{\erf}{{\rm erf}}

\newcommand{\sst}{\scriptstyle }
\newcommand{\xparen}{\mbox{\small$(\bfx)$}}
\newcommand{\hojz}{H_{0j}\mbox{\small$(\bfz)$}}
\newcommand{\Hozj}{H_{0,j}\mbox{\small$(\bfz_j)$}}
\newcommand{\smallmath}[1]{{\scriptstyle #1}}
\newcommand{\Hoz}[1]{H_0\mbox{\small$(#1)$}}
\newcommand{\Hozp}[1]{H_0^\prime\mbox{\small$(#1)$}}
\newcommand{\Hozpp}[1]{H_0^{\prime\prime}\mbox{\small$(#1)$}}
\newcommand{\hoz}{\Hoz{\bfz}}
\newcommand{\hooz}{\Hozp{\bfz}}
\newcommand{\hoooz}{\Hozpp{\bfz}}
\newcommand{\smJ}{{\scriptscriptstyle \! J}}
\newcommand{\smK}{{\scriptscriptstyle \! K}}

\newcommand{\erfc}{{\rm erfc}}
\newcommand{\bitem}{\begin{itemize}}
\newcommand{\dsum}{{ \displaystyle \sum}}
\newcommand{\eitem}{\end{itemize}}
\newcommand{\benum}{\begin{enumerate}}
\newcommand{\eenum}{\end{enumerate}}
\newcommand{\bdm}{\begin{displaymath}}
\newcommand{\bfzro}{{\underline{\bf 0}}}
\newcommand{\bfone}{{\underline{\bf 1}}}
\newcommand{\edm}{\end{displaymath}}
\newcommand{\beq}{\begin{equation}}
\newcommand{\bea}{\begin{eqnarray}}
\newcommand{\eea}{\end{eqnarray}}
\newcommand{\cali}{ {\cal \bf I}}
\newcommand{\caln}{ {\cal \bf N}}
\newcommand{\barray}{\begin{displaymath} \begin{array}{rcl}}
\newcommand{\earray}{\end{array}\end{displaymath}}
\newcommand{\eeq}{\end{equation}}
\newcommand{\btheta}{\mbox{\boldmath $\theta$}}
\newcommand{\bTheta}{\mbox{\boldmath $\Theta$}}
\newcommand{\blam}{\mbox{\boldmath $\Lambda$}}
\newcommand{\beps}{\mbox{\boldmath $\epsilon$}}
\newcommand{\bdelta}{\mbox{\boldmath $\delta$}}
\newcommand{\bgamma}{\mbox{\boldmath $\gamma$}}
\newcommand{\balpha}{\mbox{\boldmath $\alpha$}}
\newcommand{\bbeta}{\mbox{\boldmath $\beta$}}
\newcommand{\balphascript}{\mbox{\boldmath ${\scriptstyle \alpha}$}}
\newcommand{\bbetascript}{\mbox{\boldmath ${\scriptstyle \beta}$}}
\newcommand{\bLambda}{\mbox{\boldmath $\Lambda$}}
\newcommand{\bDelta}{\mbox{\boldmath $\Delta$}}
\newcommand{\bomega}{\mbox{\boldmath $\omega$}}
\newcommand{\bOmega}{\mbox{\boldmath $\Omega$}}
\newcommand{\blambda}{\mbox{\boldmath $\lambda$}}
\newcommand{\bphi}{\mbox{\boldmath $\phi$}}
\newcommand{\bpi}{\mbox{\boldmath $\pi$}}
\newcommand{\bnu}{\mbox{\boldmath $\nu$}}
\newcommand{\brho}{\mbox{\boldmath $\rho$}}
\newcommand{\bmu}{\mbox{\boldmath $\mu$}}
\newcommand{\sigi}{\mbox{\boldmath $\Sigma$}_i}
\newcommand{\bfu}{{\bf u}}
\newcommand{\bfx}{{\bf x}}
\newcommand{\bfb}{{\bf b}}
\newcommand{\bfk}{{\bf k}}
\newcommand{\bfc}{{\bf c}}
\newcommand{\bfv}{{\bf v}}
\newcommand{\bfn}{{\bf n}}
\newcommand{\bfK}{{\bf K}}
\newcommand{\bfh}{{\bf h}}
\newcommand{\bff}{{\bf f}}
\newcommand{\bfg}{{\bf g}}
\newcommand{\bfe}{{\bf e}}
\newcommand{\bfr}{{\bf r}}
\newcommand{\bfw}{{\bf w}}
\newcommand{\calX}{{\cal X}}
\newcommand{\calZ}{{\cal Z}}
\newcommand{\bb}{{\bf b}}
\newcommand{\bfy}{{\bf y}}
\newcommand{\bfz}{{\bf z}}
\newcommand{\bfs}{{\bf s}}
\newcommand{\bfa}{{\bf a}}
\newcommand{\bfA}{{\bf A}}
\newcommand{\bfB}{{\bf B}}
\newcommand{\bfV}{{\bf V}}
\newcommand{\bfZ}{{\bf Z}}
\newcommand{\bfH}{{\bf H}}
\newcommand{\bfX}{{\bf X}}
\newcommand{\bfR}{{\bf R}}
\newcommand{\bfF}{{\bf F}}
\newcommand{\bfS}{{\bf S}}
\newcommand{\bfC}{{\bf C}}
\newcommand{\bfI}{{\bf I}}
\newcommand{\bfO}{{\bf O}}
\newcommand{\bfU}{{\bf U}}
\newcommand{\bfD}{{\bf D}}
\newcommand{\bfY}{{\bf Y}}
\newcommand{\bSig}{{\bf \Sigma}}
\newcommand{\test}{\stackrel{<}{>}}
\newcommand{\zmk}{{\bf Z}_{m,k}}
\newcommand{\zlk}{{\bf Z}_{l,k}}
\newcommand{\zm}{{\bf Z}_{m}}
\newcommand{\ssq}{\sigma^{2}}
\newcommand{\dint}{{\displaystyle \int}}
\newcommand{\ds}{\displaystyle }
\newtheorem{theorem}{Theorem}
\newcommand{\postscript}[2]{ \begin{center}
    \includegraphics*[width=3.5in,height=#1]{#2.eps}
    \end{center} }